\pdfoutput=1

\documentclass[11pt]{article}

\usepackage{ACL2023}

\usepackage{times}
\usepackage{latexsym}

\usepackage[T1]{fontenc}

\usepackage[utf8]{inputenc}

\usepackage{microtype}

\usepackage{inconsolata}

\usepackage{hyperref}
\usepackage{booktabs}
\usepackage{tabularx}
\usepackage{bbding}
\usepackage{graphicx}
\usepackage{subcaption}
\usepackage{array}
\usepackage{thanks-nostar}

\newcolumntype{C}{>{\centering\arraybackslash}X}

\title{A Survey of Vision-Language Pre-training\\ from the Lens of Multimodal Machine Translation}

\author{Jeremy Gwinnup$^{1,2}$,  Kevin Duh$^1$ \\
$^1$Johns Hopkins University, $^2$Air Force Research Laboratory \\
{\tt jeremy.gwinnup.1@us.af.mil, kevinduh@cs.jhu.edu}\\
}

\begin{document}
\maketitle
\begin{abstract}
Large language models such as BERT and the GPT series started a paradigm shift that calls for building general-purpose models via pre-training on large datasets, followed by fine-tuning on task-specific datasets. 
There is now a plethora of large pre-trained models for Natural Language Processing and Computer Vision.  
Recently, we have seen rapid developments in the joint Vision-Language space as well, where pre-trained models such as CLIP \cite{radford-clip-2021} have demonstrated improvements in downstream tasks like image captioning and visual question answering.  
However, surprisingly there is comparatively little work on exploring these models for the task of multimodal machine translation, where the goal is to leverage image/video modality in text-to-text translation.  
To fill this gap, this paper surveys the landscape of language-and-vision pre-training from the lens of multimodal machine translation. 
We summarize the common architectures, pre-training objectives, and datasets from literature and conjecture what further is needed to make progress on multimodal machine translation. 
\end{abstract}

\section{Introduction}

Improving translation quality of human language by machines is a continuous goal, usually focusing on collection of parallel text in the source and target languages towards creating a more capable model. When looking at the problem of translation when presented with additional information in the form of an image caption or a video, there is a natural inclination to use the information present in the visual modality to provide context when translating the accompanying text.

Various shared tasks at the Conference for Machine Translation \cite{specia-etal-2016-shared,elliott-etal-2017-findings} and the Workshop on Asian Translation \cite{emnlp-2019-asian,nakazawa-etal-2020-overview,nakazawa-etal-2021-overview,nakazawa-etal-2022-overview} were held to encourage research into this question to varying degrees of success when largely working with small, parallel corpora paired with imagery such as Multi30k \cite{elliott-etal-2016-multi30k}. \citet{Sulubacak2020} survey a number of methods that train MMT systems from scratch.

With the recent development of Vision-Language models such as CLIP \cite{radford-clip-2021}, it seems like a natural next step to leverage these large, multimodal models to provide context in order to improve translation quality. Surveys on pre-trained Vision-Language models \cite{https://doi.org/10.48550/arxiv.2202.09061,https://doi.org/10.48550/arxiv.2202.10936,https://doi.org/10.48550/arxiv.2302.10035} outline various forms of models that may aid in using vision information in machine translation.

In the following sections, we will discuss challenges to Multimodal Machine Translation (MMT), including strong versus weak grounding, datasets available from which to train systems, various Vision-Language models and architectures that can aid in this task and datasets used to train these models and MMT systems. Lastly, we present a method to roughly evaluate a datasets relative strength of grounding.

\begin{figure*}[htbp]
    \centering
    \includegraphics[width=\textwidth, angle=0]{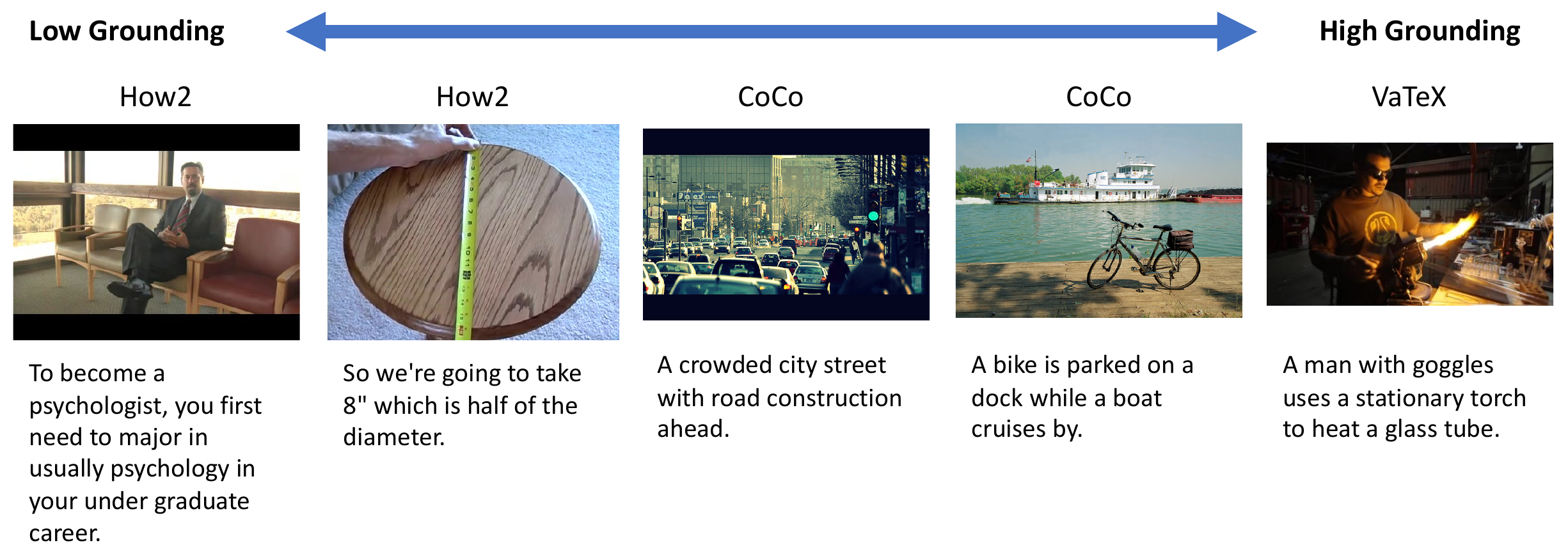}
    \caption{Examples from CoCo, How2 and VaTeX demonstrating varying degrees of grounding from low to high.}
    \label{fig:groundingex}
\end{figure*}

\section{Multimodal Machine Translation}

\paragraph{Task:} Multimodal Machine Translation (MMT) is conventionally defined as the challenge of improving text-to-text translation with visual signals, which are either images or videos. 
Given training data the consists of the triplets source sentence $x$, target translation $y$, and visual signal $v$, we train a MMT model:

\begin{equation}
P(y | x, v) 
\end{equation}

At test time, we are given $x_{test}$ and $v_{test}$ as inputs, and wish to generate a translation prediction $\hat{y}_{test}$ which is ideally close to the reference $y_{test}$.
A MMT model is considered successful if it outperforms the baseline $P(y|x)$ which does not utilize the visual signal. 

The survey paper by \citep {Sulubacak2020} summarizes MMT research up to 2019, focusing on MMT models that are trained from scratch rather than from pre-trained models.

\paragraph{Strong vs. Weak Grounding:} Importantly, we must clarify the extent to which $x$ and $v$ are related. 
In the popular WMT Multimodal shared task \cite{specia-etal-2016-shared,elliott-etal-2017-findings}, $x$ is a caption of the image $v$; similarly in VaTeX, $x$ is crowdsource-generated caption of the video $v$. 
In other words, $x$ is an accurate description of what is happening in the $v$. 
In this case, we can say that $x$ is grounded in $v$, and can expect $v$ to give useful information for translation of $x$.

On the other end of the spectrum, imagine a movie's video $v$ and its associated subtitle $x$. 
The grounding may be much weaker. For example, the video may show the faces of two characters discussing some event, in which case $x$ has no relation to $v$. A translator for these subtitles probably do not need to reference the video, and can work soley based on the text. 
In the middle of the spectrum are datasets like How2 \cite{sanabria18how2}, where speakers describe how to perform something, so there is likely grounding for some of the objects or actions but not the entire $x$. 

One may argue whether strongly-grounded MMT datasets are realistic translation problems that occur naturally. Even if one's opinion is negative, one can consider these MMT datasets as easier proxy problems to study vision-language. 

A different question is whether the datasets for V-L pre-training should be strongly-grounded. It turns out the majority are strongly-grounded like image captions due to the way the datasets are prepared. How would a strongly-grounded pre-training dataset work for a weakly-grounded MMT task? The answer is not obvious.

As a quick case study examining dataset grounding, we look at the CoCo Image Captioning, the How2 video instruction and VaTeX video captioning datasets. Each of these datasets has different aims which effect the levels of grounding between text and imagery or video segments. For instance, we expect CoCo and VaTeX datasets to have stronger grounding as whole as the dataset annotator's task is to describe the image or actions in a video clip. How2 presents a different challenge as the text segments are the utterances of the narrator of the video describing the task they are demonstrating and this may not fully represent what is currently shown in the video.

Even with these differences in annotation objectives, segments within a dataset may demonstrate different levels of grounding. Figure \ref{fig:groundingex} shows examples with varying degree of grounding on a spectrum from low to high. While we have no formal measure of grounding for these examples, we can still roughly categorize a difference on how closely a caption relates to the visual element. We do note that samples within each dataset  can vary on this grounding spectrum, but on the whole, datasets that focus on image and video captioning (versus narration) tend to demonstrate stronger grounding.

\section{Datasets} 
\label{sec:datasets}

Traditionally, MMT systems were trained on small standalone image and multiple-language text corpora such as Multi30k \cite{elliott-etal-2016-multi30k}. The size of these corpora were typically limited by the effort needed to hand-annotate images and then provide translations of the annotations. Later corpora such as How2 \cite{sanabria18how2}, VaTeX \cite{https://doi.org/10.48550/arxiv.1904.03493} and QED/Amara \cite{abdelali-etal-2014-amara} introduced larger training data (again human annotated) for a visual modality and parallel text segments in two or more languages. Additional video-text datasets such as VISA  \cite{li-etal-2022-visa} have been constructed to specifically examine effects of video features to help resolve ambiguous word choices in translation.

The common feature of the above datasets is they are collected in a strongly-supervised manner where humans are employed to guarantee that the video or image and text annotations (in multiple languages) correlate to each other, even if not in direct captioning setting. This unfortunately limits the size of these datasets due to the costs inherent in human annotation.

\begin{figure}[h]
	\centering
	\includegraphics[scale=0.335]{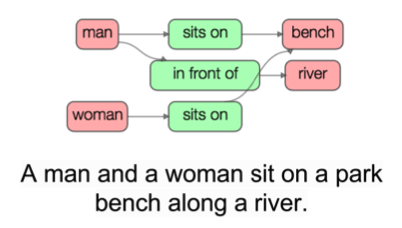}
	\caption{Example scene graph from Conceptual Captions dataset \cite{sharma-etal-2018-conceptual}}
	\label{fig:scenegraphexample}
\end{figure}

If we are not as concerned with providing data in a second language, we can reduce the amount human labor needed to create dataset. Image captioning datasets like COCO \cite{https://doi.org/10.48550/arxiv.1405.0312} provide multiple captions for over 120k images, yielding more captions than the MMT-centric resources listed above. Additional work in representing the structure of objects in a scene via a graph was the focus of the Visual Genome \cite{https://doi.org/10.48550/arxiv.1602.07332} dataset. An example scene graph is shown in Figure \ref{fig:scenegraphexample} demonstrating the relationship between objects in a scene. This approach does require more hands-on annotation than a traditional video captioning task and as such contains fewer samples than a captioning-only dataset.

The MSR-VTT \cite{xu2016msr-vtt} dataset collects 10k video clips in 20 categories towards research in open-domain video captioning. Again, these samples are hand-annotated using Mechanical Turk, which limits the potential size of the collected data.

More modern approaches rely on unsupervised methods to harvest images and alt-text from web pages in order to construct massive corpora of image-text pairs. Conceptual Captions \cite{sharma-etal-2018-conceptual}, LAION5B \cite{https://doi.org/10.48550/arxiv.2210.08402} and WebLI \cite{https://doi.org/10.48550/arxiv.2209.06794} datasets have successfully collected millions and \textit{billions} of pairs, but state that a series of filtering steps are needed to ensure the resulting dataset is not too noisy. Similarly, the YFCC100M \cite{Thomee_2016} dataset examines images and videos and uses classifiers to determine their content, though captions or text content may not be present in all samples.

Table \ref{tab:datasets} provides a list of datasets for both traditional MMT datasets comprised of image/video, source language and target language segments and larger text-image pair corpora used to train models such as CLIP and Vision Transformers. We denote the strength of grounding of the vision and language components of each dataset with asterisks with one representing relatively weak grounding and three representing strong grounding. We recall earlier discussion that image and video captioning datasets tend to have strong grounding where datasets focusing on instructional videos such as How2 will have weaker grounding due to the demonstrator's narration not always directly describing the current video scene.

\begin{table*}[h]
    \centering
    \begin{tabularx}{\linewidth}{l X X X}
    	\toprule
    	Dataset & $x$-$v$ Grounding & Size & Language \\
   	\midrule
    	\multicolumn{4}{c}{MMT} \\
	\midrule
	Multi30k & image captions ($****$) & 30k images, 30k captions & cs, de, en, fr \\
	QED (Amara) & subtitles of lectures ($*$) & 23.1k video clips,  8k-335k segments & 20 languages \\
	How2 & subtitles of instructional videos ($**$)& 13k video clips, 189k segments& en, pt \\
	VaTeX & video captions ($****$) & 41k video clips, 206k captions & en, zh \\
	VISA & video captions  ($****$) & 40k video clips + captions& en, ja \\
	\midrule
	\multicolumn{4}{c}{Vision-Language} \\
	\midrule
	MS-COCO & image captions ($****$) & 120k images, 600k captions & en \\
	Visual Genome & scene graphs ($****$) & 108k images, 109k scene graphs &  en\\
	MSR-VTT & video annotations ($****$) & 10k video clips, 200k annotations & en \\	
	Conceptual Captions & image desc. ($****$)& 3.3m image caption pairs&  en \\
	YFCC100M & image classification ($**$) & 99.2m image, 800k video clips & indeterminate \\ 
	LAION 5B & image + alt text  ($***$) & 5.85B image text pairs & 109 languages\\
	WebLI & image  + alt text  ($***$) & 10B image text pairs & 100+ languages \\
	\bottomrule
    \end{tabularx}
    \caption{Datasets for MMT and V-L. ($****$) indicates our opinion on the relative strength of grounding, more asterisks means strong $x$-$v$ relationship.}
    \label{tab:datasets}
\end{table*}

\section{Pre-training Objectives}

With the problem of MMT defined and an understanding of the available datasets, we next categorize three broad classes of pre-training objectives seen in the literature: Masking, Matching, and Dataset-specific. For each of the methods, we briefly discuss common ``building blocks'' used by the models we review later in Section \ref{sec:modelarch}.

\subsection{Masking}

As with the text-only Bidirectional Encoder Representations from Transformers (BERT) \cite{devlin-etal-2019-bert} model, a common training objective for certain vision-language models is Masking, which removes tokens from a sequence and asks the model to predict those missing tokens. The Vision Transformer \cite{https://doi.org/10.48550/arxiv.2010.11929} (ViT) extends this approach to ingest image patches as tokens, concatenating these image tokens with the corresponding text tokens.  \citet{he2021masked} state that masking autoencoders (including ViT) are scalable self-supervised learners due to this approach.
The transformer-only ViT approaches meet or exceed performance of recent CNN methods when trained with sufficient data in large-network configurations.

\subsection{Matching}

The next pre-training objective we examine is masking, where we wish to minimize the embedding-space distance between text and visual pairs. Traditional image-processing centric methods using CNNs have realized success in captioning a presented image, but typically require a very deep stack of convolutional layers and lengthy training times to accommodate these deep networks. 
Recent approaches such as Contrastive Language-Image Pretraining (CLIP) \cite{radford-clip-2021} and ConVIRT \cite{https://doi.org/10.48550/arxiv.1904.05342}  scale a simpler version of this matching pre-training task up to much larger amounts of data to train a more-generally capable model not limited to a predefined set of classes.
 In order to reach this larger data requirement, these two models train in a series of batches in a matrix orientation where the correct examples are also compares with all of the other incorrect pairings. The model then learns how to minimize the contrastive loss in order to favor the correct pairings.
By training a paired image and text encoder to predict which images are paired with the corresponding text in the dataset, we can use resulting model to label an unseen image in a zero-shot manner not requiring additional fine-tuning.

\subsection{Dataset-specific Supervised Objective}

The last pre-training objective we discuss is specific to the dataset used. As shown in Section \ref{sec:datasets}, these datasets tend to be Visual Captioning (VC) or Visual Question Answer (VQA). 
For VC, the objective is to use image-only inputs to generate sentence captions. For VQA, one approach is to use the text prediction capability of a LLM to effectively generate an answer from a text sequence containing a question. In the context of Vision-Language models, image tokens can form part of this question sequence (along with question-based text) in order to perform VQA. 
These objectives are supervised, but this is not a limitation since large quantities of VC/VQA dataset have been created.

\subsection{Discussion}

With the naming of the three above pre-training objectives, we can now take a look at how these strategies can help improve multimodal machine translation. The most obvious approach to consider is to use large, pretrained joint-encoder vision-language models to serve as the front-end of a translation model, however this is easier said than done. The only work that we are aware of in applying large pre-training to MMT is \citet{caglayan-etal-2021-cross}, who examine pre-training with a 3-way image and bilingual text corpus building on the approach of \citet{lample2019crosslingual}, but run into the common problem of data availability. They address this issue by using a text-only translation model to translate the English text of the Conceptual Captions \cite{sharma-etal-2018-conceptual} corpus to German, synthesizing data \cite{sennrich-etal-2016-improving} to facilitate building the integrated translation model.

\section{Model Architectures}
\label{sec:modelarch}

Next we match a series of models to one or more of the above pre-training objectives and go into further detail on how these models accomplish them.

We begin by examining which modalities the models we discuss are capable of processing, listed in Table \ref{tab:modelinputs}. The majority of the models we examine are text-and-image models, with exception of VideoBERT \cite{https://doi.org/10.48550/arxiv.1904.01766} working on videos and  Flamingo \cite{https://doi.org/10.48550/arxiv.2204.14198} working on both images and video.

\begin{table}[ht]
	\centering
	\begin{tabular}{l c c c }
	\toprule
	& \multicolumn{3}{c}{input modality} \\
	model & text & image & video \\
	\midrule
	ConVIRT & X & X & \\
	CLIP & X & X & \\
	Flamingo & X & X & X \\
	VideoBERT & X & & X \\
	ICMLM & X & X & \\
	GiT & X & X & \\
	BeiT & X & X & \\
	BLIP-2 & X & X & \\
	ALIGN & X & X & \\
	CoCa & X & X & \\
	\bottomrule
	\end{tabular}
	\caption{Model Input Modalities}
	\label{tab:modelinputs}
\end{table}

An overview of various models' encoder and decoder architectures and modality fusion strategies are listed in Table \ref{tab:modelencdec}. We note that the majority of models discussed use a transformer like ViT as the image encoder, CNNs are also a popular choice. For the text encoder components, the overwhelming choice is BERT-style architectures.

\begin{table*}[ht]
	\centering
	\begin{tabularx}{\linewidth}{l X X X}
	\toprule
	model & vision encoder & text encoder & fusion  \\
	\midrule
	ConVIRT & CNN + classifier& ClinicalBERT \cite{https://doi.org/10.48550/arxiv.1904.05342} & Max agreement between bidirectional loss\\
	CLIP&ResNet50 / ViT-L & GPT-2 \cite{radford2019language} & Transformer-style QKV\\
	Flamingo & Perceiver & Chinchilla & Interleaved gated vision layers in frozen text model\\
	VideoBERT & BERT w/ Visual Tokens& BERT & Early fusion with text and vision tokens\\
	ICMLM & CNN & BERT& Proxy task to learn visual representations over image-caption pairs\\
	GiT & Swin-like vision-to-text transformer (no obj det)& BERT & Trained with LM task. Scaling data size helps\\
	BeiT-3& VLMo vision expert FFN & VLMo text expert &Shared Multi-head self-attn, modality routing\\
	BLIP-2 & ViT with frozen image encoder & BERT with frozen text encoder & Query Transformer\\
	ALIGN & EfficientNet  CNN w/ global pooling & BERT & Modality Contrastive loss (norm softmax); fully connected layer w/ linear activation\\
	CoCa & ViT / CNN& BERT& contrastive loss\\
	PALI & ViT-G/e & T5-XXL &  ViT and text feed transformer encoder\\
	\bottomrule
	\end{tabularx}
	\caption{Model Encoder Decoder Architectures}
	\label{tab:modelencdec}
\end{table*}

The real difference between these models is how the information from the modalities is combined. As described above, CLIP and ConVIRT aim to minimize the contrastive loss between batches of image and text pairs. 
ViT models sample patches of the image input and concatenate the resulting tokens to the text tokens.

Similar to ViT, the Swin Transformer \cite{liu2021swin} employs a hierarchical transformer model using shifted windows of varying sizes to build a hierarchy at different scales.

VirTex \cite{desai-2020-virtex} takes a different approach, jointly training a CNN and a transformer on image-caption pairs and then transfer the learned CNN to downstreams tasks such as object detection.
ICMLM \cite{https://doi.org/10.48550/arxiv.2008.01392} also uses a CNN for image features and BERT, but utilize a proxy tasks such as extracting image tags (such as part of speech) from text captions, then uses masking to learn the visual representations.  %
 PALI \cite{https://doi.org/10.48550/arxiv.2209.06794}  works at a large scale, training on over 10B image text pairs using a large ViT and BERT-like T5-XXL \cite{raffel2020exploring} text model.
 BEiT-3\cite{https://doi.org/10.48550/arxiv.2208.10442}  treats the image modality as a foreign language.  The key to this approach is using VLMo \cite{https://doi.org/10.48550/arxiv.2111.02358} to route sequences to the appropriate modality ``expert''.
 
Flamingo \cite{https://doi.org/10.48550/arxiv.2204.14198}  does things differently in order to scale the model to larger data. A Perceiver \cite{https://doi.org/10.48550/arxiv.2103.03206}, a transformer model with asymmetric attention mechanism is used as the vision encoder, Chinchilla \cite{https://doi.org/10.48550/arxiv.2203.15556},  a compute-efficient LLM trained on 1.4 billion tokens, is used as the text encoder. The layers of the text model are frozen (held constant) and vision layers are inserted at various depths to integrate visual information into the network. Freezing the text layers ensures that pre-trained information is not lost with network updates during training.
The Generative Image-to-text Transformer (GiT) \cite{https://doi.org/10.48550/arxiv.2205.14100} uses a similar approach (though without interleaving) and does not freeze the text model.
In another permutation of vision-language model combination, BLIP-2 \cite{https://doi.org/10.48550/arxiv.2301.12597}  combines off-the-shelf separate vision and language models with a lightweight querying transformer with a two-stage pre-training process. Here, both models are frozen, with the learned query transformer acting as a gatekeeper feeding the processed input visual feature to the text model resulting in the final text output.

ALIGN \cite{https://doi.org/10.48550/arxiv.2102.05918} uses frequency-based sampling on noisy text data. While largely following the method used in collecting the Conceptual Captions dataset, the simpler frequency-based sampling takes the place of the more complex filtering  in \citet{sharma-etal-2018-conceptual}, resulting in a much larger dataset. 
1B training samples. calc image-to-text loss and text-to-image loss. EfficientNet image encoder, BERT text encoder.
CoCa \cite{https://doi.org/10.48550/arxiv.2205.01917} is an image-text encoder-decoder model jointly trained with contrastive loss and generative loss. This model uses the popular combination of ViT for image encoding and BERT for text encoding. 

One of the few models discussed working with video instead of images, VideoBERT \cite{https://doi.org/10.48550/arxiv.1904.01766}  apply BERT's approach to video frames, by extracting ``visual words'' using CNNs to sample image features at a 20 frames-per-second (fps) rate, then averaging those features for the frames that correspond to the segment's spoken text. These extracted features are added as additional tokens to the text during training.

\begin{table*}[h!]
	\centering
	\begin{tabular}{l c c c c c c}
		\toprule
		Dataset  & Samples & Reference & Reference 2 & Neighbor & In-Domain & Out-Domain \\
		\midrule
		CoCo & 82,783 & 48.69\% & 51.17\% &  & 0.11\% & 0.03\% \\
		CoCo Diffusion & 82,783 & 85.75\% & 14.13\% &  & 0.07\% & 0.04\% \\
		How2 Clip@ 2 sec.& 126,213 & 51.23\% &  & 33.42\% & 9.04\% & 5.62\% \\
		How2 Diffusion &188, 949 & 82.02\% & & 12.80\% & 2.75\% & 3.06\% \\ 
		VaTex Clip@ 2 sec. & 23,740 & 49.31\% & 49.57\% & & 0.88\% & 0.24\%  \\
		VaTex Diffusion & 23,740 & 45.27\% & 53.83\% & & 0.53\% & 0.37\%  \\
		\bottomrule
	\end{tabular}
	\caption{CLIP Label class percentages for various corpora.}
	\label{tab:clipscorebins}
\end{table*}

\section{Analysis}

Our goal in this survey is to understand whether the large body of literature on V-L models can potentially have impact on the task of MMT. In Figure~\ref{fig:groundingex}, we've identified grounding strength as an important factor influencing whether visual input complements the text modality. We argued that samples in MMT tasks like How2 generally have low grounding, where translations are loosely related to the video, while those of COCO and VaTeX have high grounding because they are constructed like translations of captioning datasets. V-L models are generally trained on datasets with high grounding according to our definition, so it would be interesting to analyze whether V-L models behaves fundamentally different for How2 vs COCO/VaTeX.

\subsection{Measuring Grounding as a Zero-Shot Classification Task}

We apply V-L models to MMT dataset in the following way: Rather than building a V-L model for translation, we use V-L models to exploit visual input to ``re-rank" a set of sentences, one of which is the reference. We compare V-L models' relative performance on this reranking task across MMT datasets to understand the potential fit for this problem.
In particular, we look to use the zero-shot classification capabilities of CLIP to evaluate an image or still clip and text reference to see which of four provided labels scores closest to the visual component:

 \begin{itemize}
	\itemsep0em 
	\item The reference segment
	\item A randomly sampled neighboring caption from the same caption set or the same video or image.
	\item A randomly sampled text segment from the training portion of the dataset (Considered to be in-domain data).
	\item A randomly sampled text segment from an unrelated corpus (Considered to be out-of-domain data).
\end{itemize}

We argue that a dataset is highly grounded when the CLIP score is highest with the reference segment, meaning that the text and image pair are close in the models joint embedding space.

We next perform this analysis on the CoCo, How2 and VaTeX datasets. For all three datasets, we use the first 10 million lines from English portion of the Europarl \cite{koehn-2005-europarl}, v10, to represent out-of-domain data; the intuition is that all previously considered datasets in this work are markedly different in content than Europarl's parliamentary proceedings.

For the CoCo and VaTex datasets, since each image or video has multiple captions, we consider the first caption to be reference and the additional captions for that image are randomly sampled as a second reference for comparison. %
For the How2 dataset, each video has multiple spoken segments; the reference line from the training corpus is match to the originating video and other text segments from the same video are used as the pool of neighbors. We use the the collection of reference captions as the pool from which to sample in-domain labels. 

We also create a ``mirror-corpus'' of each of the three corpora we examine by using the English caption or segment as a prompt supplied to a latent diffusion model trained to output images from text prompts such as Stable Diffusion \cite{https://doi.org/10.48550/arxiv.2112.10752}. These additional corpora provide another comparison point with the same four-class arrangement.

In Table \ref{tab:clipscorebins} we show which percentage of the above four label classes score highest for each condition. We consider a dataset to be strongly grounded if the reference label scores the highest.  

We notice that CoCo and VaTeX datasets strongly favor either of the two reference labels, intuitively this makes sense as both datasets are constructed for captioning.

For the How2 dataset, we notice that the reference label is strongest about 51\% of the time, with the neighbor class is strongest about 33\% of the time. This suggests the text components relate more to the video at large and not the specific contents of the video for the time segment corresponding to the reference text. When considering the How2 scores for diffusion imagery, we see that the reference label is selected the vast majority of the time, similar to CoCo's diffusion imagery results. This is expected since that reference is the prompt supplied to the diffusion model, which also uses CLIP as a text encoder.

\section{Conclusion}

In conclusion, we discuss some of the challenges facing effective MMT, describe datasets currently used to train MMT and V-L models, and then review some architectures of these models, noting commonalities in many approaches. Finally we discuss a zero-shot classification method to determine the relative groundedness of a multimodal dataset and test the technique on three corpora.

\thanksnostar{
Opinions, interpretations, conclusions and recommendations are those of the authors and are not necessarily endorsed by the United States Government. Cleared for public release on 05 June 2023. Originator reference number RH-23-124431. Case number AFRL-2023-2732.
}

\bibliography{custom}
\bibliographystyle{acl_natbib}

\end{document}